INPUT THE TYPE OF MANUSCRIPT

# A Monkey Swing Counting Algorithm Based on Object Detection

Hao Chen†, Zhe-Ming Lu† *, and Jie Liu ‡

**SUMMARY** This paper focuses on proposing a deep learning-based monkey swing counting algorithm. Nowadays, there are very few papers on monkey detection, and even fewer papers on monkey swing counting. This research focuses on this gap and attempts to count the number of monkeys swinging their heads by deep learning. This paper further extends the traditional target detection algorithm. By analyzing the results of object detection, we localize the monkey's actions over a period of time. This paper analyzes the task of counting monkey head swings, and proposes the standard that accurately describes a monkey swinging its head. Under the guidance of this standard, the head-swing count in 50 monkey movement videos in this paper has achieved 94%.
*key words:* Monkey Detection, Object Detection, Head Swing Counting, YOLO.

## 1. Introduction

Counting the number of times monkeys shake their heads is mainly in the field of biomedicine [1]. In these fields, monkeys are important experimental subjects, and counting the number of monkeys shaking their heads is of great significance for judging abnormal monkey behaviors. Accurate head swing counting is an important indicator to verify the experimental results. In the past, the number of monkey swings was usually counted manually. This method is very accurate, but once the length of the video is large or there are many videos, it takes almost the same amount of time as the video to count. Manual counting consumes a lot of manpower. This paper focuses on this contradiction and hopes to propose a method based on deep learning, which can automatically detect monkeys in a video, locate them, and then count the number of times the monkeys swing their heads in a video [7].

In recent years, as one of the three major tasks in computer vision, object detection has developed rapidly, and a large number of excellent works have emerged, such as R-CNN [14], SSD [8], and YOLO [10]. R-CNN uses two parts, one extracting about 2000 regions through the RPN module, the other judging whether these areas contain the target through a classifier. YOLO and SSD output all information including the target frame and the probability that the target belongs to different categories through a network. The detection results of the R-CNN algorithm are more accurate, but the training process leads to high training complexity. The YOLO algorithm is simple enough and has a low training cost, which is favored by the industry, but the detection accuracy is lower than R-CNN. Considering the needs of this project in applications, we mainly use the YOLO algorithm as the detection algorithm.

This paper implements the monkey swing counting algorithm by extending the YOLO algorithm. Considering that this project only counts the number of monkeys shaking their heads, we trained the model with a set of monkey picture labeling heads. First, the monkey head is detected by the YOLO algorithm, and we obtain the coordinates of the bounding box [10]. Then, according to the bounding box of the monkey head, we get the positioning of the monkey head in the picture. Through continuous experiments, we get the factors that affect the swing count. Finally, we obtained the basis for judging the monkey's head swing and the parameters that accurately describe the monkey's head swing.

The second paragraph of this paper mainly describes the relevant algorithm of the monkey swing counting; the third paragraph mainly introduces the monkey swing counting method; the fourth paragraph introduces experiments and results; the fifth paragraph summarizes the full text.

*Main contributions*:
Combining target detection algorithm with biological action recognition to realize monkey swing counting based on deep learning;
Exploring the movements of the monkeys in swinging their heads, and putting forward the behavioral standards for accurately describing the monkeys' head-swinging;
Can be extended to other fields of biometric action recognition;

## 2. Related Work

2.1 Monkey Detection Algorithm

Numerous teams will begin working on monkey detection in 2020 [1]. The strategy was only employed in agriculture at the time since the academic intersection was not as great. Even while the team has started using deep learning to identify monkeys, its use is still restricted to just that and does not delve deeply into the target's action description. The recognition of human key points has increased along with deep learning, and some researchers have even adapted key point detection to monkeys [2] [3] [4] [5] [6]. Through the detection of key points of monkeys, we have been able to capture the movements of monkeys, but there is still no precise quantitative indicator. In this paper, we combine the monkey detection and morphology, and give the basis for accurately describing the monkey head swing.

†The author is in School of Aeronautics and Astronautics Zhejiang University, Hangzhou 310027, P.R. China.
‡The author is in Centre for Excellence in Brain Science and Intelligence Technology, Chinese Academy of Science.
*Corresponding author: zheminglu@zju.edu.cn.

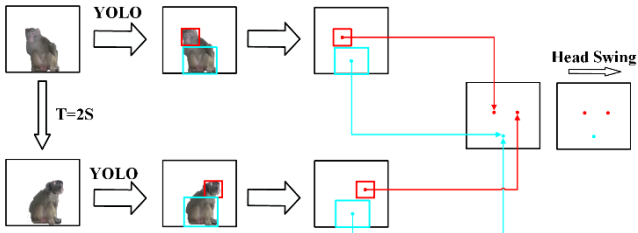

**Fig. 1** Compare the changes of the monkey's head position before and after 50 frames to determine whether the monkey swings its head.

2.2 Object Detection Algorithms

As one of the three major tasks in deep learning, object detection have always attracted a lot of attention. With the advent of the ResNet [15], deep learning has developed rapidly. We propose a model that performs well in the object detection, R-CNN [14]. This model still affects two important tasks in deep learning, object detection and segmentation. However, the model needs to be trained twice, which greatly increases the complexity of the model.

In order to further simplify the training complexity of the model, SSD [8] and YOLO [10] have come out one after another. They get all the information including the target bounding box and the probability that the target belongs to different classes through one model. Due to its high accuracy and efficiency, the YOLO [10] [12] [13] was soon widely recognized by the industry, and multiple versions were derived.

In recent years, in order to describe the human pose, key point detection has become more and more popular [11]. However, the related technologies are not mature enough to deal with the complex scenarios in this project. Therefore, we adopt YOLO as the core algorithm.

3. Method

3.1 Object Detection Algorithms

YOLO is a classic one-stage target detection algorithm. It is very different from R-CNN. R-CNN requires two steps in training. The model gets a lot of proposals in the first step and gets the detection results in the second step. YOLO, on the other hand, uses category and bounding box as a whole for regression. Therefore, YOLO is simpler than R-CNN.

YOLO adopts darknet53 as a backbone to extract the features of the input. The DarkNet53 is based on ResNet, which can maintain a stable gradient on the basis that the number of model layers can be continuously deepened. In order to reduce the spatial dimension of data features and speed up the training process, YOLO uses ResNet as the bottleneck to reduce the number of channels and reduce the burden of the model.

Since large-scale features are more conducive to target detection tasks, YOLO is a multi-scale model for multi-channel detection. The low-level features complement the information lost during training. The multi-scale model also enriches the feature fusion of the model in training.

The main contribution of YOLO is the loss function. The model considers the classification loss and regression loss together. YOLO uses the L2 regression function to calculate the loss function of the BBox and uses the cross entropy to calculate the classification loss, including foreground and background.

The accuracy of the target detection algorithm determines the accuracy of the monkey swing count.

3.2 Monkey Action Recognition Algorithm

The ideal situation of this project is to transplant the standard of manual counting directly into the algorithm. However, human subjective judgment is often difficult to form an effective algorithm criterion. This adds a lot of difficulty to the algorithm.

This project uses a number of different dimensions to determine whether the monkey is shaking its head. Considering that the most direct criterion used by humans when judging whether a monkey swings its head is the speed of swinging its head. Only when the speed is fast enough can we consider it to be an effective head-swing. At the same time, the head swing occurs in a small period of time, and only the time when the action occurs is short enough to be an effective head swing. In addition, in order to exclude the interference of small perturbations, we also introduce the amplitude as a criterion for the model.

Therefore, we use speed, time, and distance as the criteria. Only within a certain period of time, a swing with a fast enough speed and a large enough swing can be considered as an effective swing.

4. Experiments

4.1 Dataset

We collected a month of monkey activity videos as the dataset for this project. We split the videos into frames and screened out frames with different actions as much as possible, forming a dataset of 20,000 pictures, and randomly selecting 2,000 of them as test data.

We preprocess the dataset according to the YOLO algorithm. The videos collected in this project were in a monkey house without lights (to avoid light interference with monkeys). In the dark environment, the monkey skin and background color are very similar, which increases the difficulty of detection.

Our test data contains two sets of videos. One was collected from the videos of the activity monkeys in one day. We took the time when the monkeys were active during the day, and took 50 videos at equal intervals by taking a 1-minute video every 10 minutes. Another was collected in three days. By taking a 1-minute video every 10 minutes, we took out 320 videos at equal intervals. 50 videos test set contains frequent monkey activities; 320 videos test set contains a large number of videos of monkeys in the silent phase.

4.2 Criterion

We have designed an algorithm to compute the counting accuracy of the algorithm. The actual number of head swings is m, and the algorithm count is n. If |m-n|<=2, it is regarded as an accurate count, and if it exceeds 2, it is regarded as a counting error; for videos with head swings within 12 times, |m-n|-2<=10 (The number of fault tolerances is 10), the number of errors divided by 10 is the error rate; for videos with more than 12 head swings, |m-n|-2>10, the number of errors divided by the total number of fault tolerances is the error rate.

$$\text{score} = \begin{cases} 1 & , |m-n| \leq 2 \\ 1 - (|m-n|-2)/10 & , |m-n|-2 \leq 10 \\ 1 - (|m-n|-2)/|m-2|, & |m-n|-2 > 10 \end{cases} \quad (1)$$

4.3 Accuracy of Object Detection Algorithm

We tested the performance of three algorithms for monkey detection, including SSD, faster R-CNN, and YOLO. Considering the popularity of the YOLO family in the field of object detection, we tested the performance of YOLOv3, YOLOv4, and YOLOv5.
From the detection results, the latest YOLOv5 algorithm has the highest detection accuracy for the monkey head and body, and there is no obvious difference within the YOLOv5 series. Comprehensive analysis, we choose lightweight YOLOv5s6 as the main algorithm of this model.

**Table 1** Comparison of the mAP between different object algorithms in monkey detection.

| Algorithms | Backbone | input size | Training data | AP50 | AP50:95 |
|---|---|---|---|---|---|
| Faster R-CNN | Resnet50 | 640 | Trainval set | 95.3 | 76.4 |
| SSD | VGG16 | 640 | Trainval set | 92.2 | 71.3 |
| YOLOv3 | Darknet53 | 640 | Train set | 97.8 | 75.2 |
| YOLOv4 | CSPdarknet53 | 640 | Train set | 99.3 | 80.5 |
| YOLOv5s | Focus+CSP*5 | 640 | Train set | 99.6 | 81.1 |
| YOLOv5m | Focus+CSP*5 | 640 | Train set | 99.6 | 82.4 |
| YOLOv5l | Focus+CSP*5 | 640 | Train set | 99.6 | 81.5 |
| YOLOv5s6 | Focus+CSP*6 | 640 | Train set | 99.7 | 80.8 |
| YOLOv5m6 | Focus+CSP*6 | 640 | Train set | 99.7 | 81.7 |
| YOLOv5l6 | Focus+CSP*6 | 640 | Train set | 99.7 | 82.6 |

4.4 Results of the Swing Counting Algorithm

As shown in Table 2, our model achieves 94.23% and 84.92% on 50 videos and 320 videos. We explored the accuracy of algorithm counting in different scenarios and tried to standardize the criterion of monkey swinging from three dimensions: speed, amplitude, and distance.

**Table 2** Counting accuracy for both two test datasets

| Test dataset | Head speed | Body speed | Distance | Distance | Result |
|---|---|---|---|---|---|
| 50 videos | 50 | 8 | 50 | 2 | 94.23 |
| 320 videos | 50 | 8 | 50 | 2 | 84.92 |

Speed is an important indicator to evaluate whether the monkey swings its head or not. Only when the monkey head speed is fast enough can it be considered as an effective head swing. The speed of the monkey is represented by the difference in distance between ten frames before and after the monkey divided by the number of frames, speed=distance/10. Considering the interference of the monkey's walking on the head swing count, we also consider the body speed as an indicator.

**Table 3** The effect of monkey head swing speed on counting accuracy

| Test dataset | Head Speed | Result |
|---|---|---|
| 50 videos | 20 | 85.43 |
| | 30 | 88.72 |
| | 40 | 90.35 |
| | 50 | **91.03** |
| | 60 | 90.88 |
| | 70 | 90.56 |
| | 80 | 89.47 |
| | 90 | 87.52 |
| | 100 | 83.23 |
| 320 videos | 30 | 75.63 |
| | 40 | 78.5 |
| | 50 | **80.32** |
| | 60 | 79.85 |
| | 70 | 78.34 |

The results show that when the monkey moves a distance of 50 pixels within ten frames, it is an important basis for judging whether the monkey swings its head. Considering body movement as well, the maximum distance allowed for body movement is 8 pixels.

**Table 4** The effect of monkey body speed on counting accuracy, when head_speed=50

| Test dataset | Body speed | Head speed | Result |
|---|---|---|---|
| 50 videos | 6 | 50 | 90.34 |
| | 7 | 50 | 92.77 |
| | 8 | 50 | **93.91** |
| | 9 | 50 | 93.17 |
| | 10 | 50 | 91.89 |
| 320 videos | 6 | 50 | 81.85 |
| | 7 | 50 | 83.48 |
| | 8 | 50 | **84.07** |
| | 9 | 50 | 82.56 |
| | 10 | 50 | 80.99 |

In order to further fit the judgment of manual counting, we also consider time and distance as the criterion for judging whether the head swing. Only within 2s, if the swing amplitude of the head is more than 50 pixels, will it be considered as an effective head swing, and the experimental results also support this.

**Table 5**   The effect of monkey head swing distance on counting accuracy, when head_speed=50, body_speed=8

| Test dataset | Distance | Result |
|---|---|---|
| 50 videos | 25 | 93.98 |
|  | 50 | **94.17** |
|  | 75 | 92.70 |
| 320 videos | 25 | 84.32 |
|  | 50 | **84.77** |
|  | 75 | 84.69 |

**Table 6**   The effect of monkey head swing time on counting accuracy, when head_speed=50, body_speed=8, distance=50

| Test dataset | Time | Result |
|---|---|---|
| 50 videos | 1 | 94.21 |
|  | 2 | **94.23** |
| 320 videos | 1 | 84.87 |
|  | 2 | **84.92** |

## 5. Conclusion

This paper is an application-oriented paper on the counting of monkey head swings. We try to design an algorithm based on object detection to detect monkey head swings without human participation. Ultimately, we achieved 94% on 50 videos. There is still a certain error in the monkey positioning using target detection. Therefore, we try to locate the monkey in a better way. For example, the key point detection can directly obtain the coordinates of the monkey. However, considering the accuracy requirements of this algorithm, the algorithm with the best stability may have the best detection effect.